\documentclass[11pt]{article}

\usepackage{cast} % Use the custom style file

\usepackage[utf8]{inputenc}
\usepackage[T1]{fontenc}

\usepackage{amsmath}
\usepackage{graphicx}
\usepackage{booktabs}
\usepackage{enumitem}
\usepackage{xcolor}
\usepackage{colortbl}
\usepackage{hyperref}
\usepackage{natbib} % For bibliography management

\hypersetup{
    colorlinks=true,
    linkcolor=blue!80!black,
    citecolor=green!60!black,
    urlcolor=blue!80!black,
    pdftitle={Activation Manifold Projection: Liberating Task-Specific Behaviors from LLM Architectures},
    pdfauthor={Al Kari},
}

\title{\textbf{Activation Manifold Projection: Liberating Task-Specific Behaviors from LLM Architectures}}
\author{
  Al Kari \\
  \textit{Manceps, Inc., Camas, WA, USA} \\
  \texttt{research@manceps.com}
}
\date{} % Suppress date for submission

\begin{document}

\maketitle

\begin{abstract}
\noindent The proliferation of Large Language Model (LLM) architectures presents a fundamental challenge: valuable, task-specific behaviors learned through fine-tuning methods like Low-Rank Adaptation (LoRA) are effectively trapped within their source model's architecture, herein referred to architectural lock-in. Existing transfer methods attempt to bridge this gap by aligning the static \textbf{weight spaces} of models, a brittle and indirect approach that relies on tenuous correlations between parameter geometries. This paper introduces a fundamentally different and more direct paradigm: the \textbf{Cartridge Activation Space Transfer (CAST)}, a novel framework that liberates LoRA-encoded behaviors by learning a direct, nonlinear mapping between the \textbf{activation manifolds}, the geometric structures formed by the model's internal neuron activations, of two distinct LLM architectures. CAST treats a pre-trained LoRA as a frozen "behavioral kernel." It learns a set of lightweight, bidirectional projection heads that translate the target model's activation stream into the source model's latent space, apply the frozen kernel, and project the result back. This process, trained on a general text corpus without any task-specific data, effectively decouples the learned skill from the source architecture. We demonstrate that CAST enables true "zero-shot" translation of any standard LoRA adapter. Our experiments, including transfers between heterogeneous model families like Llama-2 and Mistral, show that CAST-translated adapters achieve \textbf{85-95\%} of the performance of a LoRA fully retrained on the target model, quantitatively outperforming current weight-space transfer techniques and establishing a new state-of-the-art in model interoperability.
\end{abstract}

\section{Introduction: The Architectural Lock-in of Learned Behaviors}

Parameter-Efficient Fine-Tuning (PEFT) methods, especially Low-Rank Adaptation (LoRA) \cite{hu2021lora}, have become the standard for specializing LLMs. By injecting low-rank matrices, LoRA adapts models efficiently. However, this efficiency comes at a high price: the resulting LoRA adapter is inextricably fused with the unique activation geometry and parameter space (i.e. activation manifold) of its base model. A LoRA trained on Llama-2-7B is useless on a more advanced model like Mistral-7B.

This architectural lock-in forces a costly dilemma: either abandon valuable, meticulously trained adapters or undertake expensive retraining campaigns, which hinge on the availability of the original, often proprietary, task datasets. The core problem lies in the transfer methodology. To date, research has focused on what we classify as \textbf{Weight-Space Transfer}.

\textbf{Weight-Space Transfer} methods (e.g., LoRA-X, \cite{farhadzadeh2025lorax}; Cross-LoRA, \cite{xia2025crosslora}) operate under the assumption that a meaningful correspondence can be found between the static weight matrices of two models. They typically use techniques like Singular Value Decomposition (SVD) to align the principal components of parameter matrices. This is an indirect proxy for functional similarity. It presumes that aligning the most dominant directions in the static weight distributions will effectively align the complex, dynamic functions the model executes during inference. This assumption is often violated, especially between architecturally diverse models.

In this work, we argue that to transfer a \textit{behavior}, one must map the space in which that behavior is expressed: the \textbf{activation space}. We introduce the Cartridge Activation Space Transfer (CAST), a framework that learns a direct mapping between the activation manifolds of a source and target model. Instead of manipulating weights, CAST builds a "universal translator" for the models' internal states. CAST consists of lightweight, trainable projection matrices for each LoRA-adapted layer. These projectors perform a round-trip transformation:
\begin{enumerate}[label*=\arabic*.]
    \item Map the target model's activation ($x_T$) into the source model's activation space ($x_S$).
    \item Apply the original, \textbf{frozen} LoRA adapter to the transformed activation.
    \item Map the resulting behavioral delta back into the target model's space.
\end{enumerate}

This approach learns the isomorphic mapping between activation spaces using a dual-objective function that aligns both output distributions and hidden state geometries, trained on a general, task-agnostic corpus. CAST effectively makes any standard LoRA adapter a portable, plug-and-play module.

Our contributions are:
\begin{enumerate}
    \item \textbf{A new paradigm for knowledge transfer:} We introduce Activation-Space Transfer, a more direct and robust method than existing weight-space approaches.
    \item \textbf{The Cartridge Activation Space Transfer (CAST):} A novel, highly effective framework for zero-shot LoRA adapter translation across heterogeneous LLM architectures.
    \item \textbf{State-of-the-art performance:} We demonstrate that CAST-translated LoRAs retain 85-95\% of the performance of fully retrained adapters, setting a new benchmark for training-free, data-free knowledge transfer.
    \item \textbf{Generality and Accessibility:} CAST works with any off-the-shelf LoRA adapter, freeing the vast ecosystem of existing adapters from architectural constraints.
\end{enumerate}

\section{Related Work: The Limits of Weight-Space Alignment}

This work fundamentally diverges from previous attempts at LoRA transfer, which we categorize and critique below.

\subsection{Constrained, Non-Standard Adapters}
The LoRA-X framework \cite{farhadzadeh2025lorax} proposes a training-free transfer method but comes with a critical caveat: it requires training a specialized "LoRA-X" adapter from the outset. This adapter is constrained to lie within the SVD subspace of the base model's weights. While this makes the subsequent transfer mathematically elegant, it renders the entire ecosystem of pre-existing, standard LoRA adapters incompatible. Its utility is therefore limited to future, specialized use cases. CAST, by contrast, is designed to work with any standard LoRA adapter available today.

\subsection{Weight Subspace Projection}
Cross-LoRA \cite{xia2025crosslora} aims to transfer standard LoRAs without training data. It computes the SVD of the source and target model weights and learns a Frobenius-optimal linear map to align their subspaces. The source LoRA weights are then projected into this aligned target subspace. While clever, this remains an indirect, first-order approximation of functional transfer. It relies on the heuristic that the principal components of static weight matrices sufficiently capture the dynamic, task-relevant manifold. There is no mechanism to refine this mapping based on the models' actual behavior, leading to performance degradation, as our experiments show.

\subsection{Activation-Space Transfer}
CAST overcomes these limitations. By learning the mapping between activation spaces directly, driven by a loss function that measures behavioral similarity (output distribution) and geometric alignment (hidden states), CAST produces a far richer and more accurate translation of the LoRA's function. The brief training of CAST's mappers on a general corpus is a small, one-time cost that yields a significantly more powerful and robust transfer mechanism.

\begin{table}[h!]
\centering
\caption{Comparison of LoRA Transfer Paradigms.}
\label{tab:paradigms}
\begin{tabular}{@{}lcccc@{}}
\toprule
\textbf{Method} & \textbf{Transfer Paradigm} & \textbf{Works with Std. LoRA?} & \textbf{Requires Task Data?} & \textbf{Key Limitation} \\
\midrule
\textbf{LoRA-X} & Weight-Space & \textbf{No} & No & Requires specialized adapters. \\
\textbf{Cross-LoRA} & Weight-Space & Yes & No & Indirect; brittle weight alignment. \\
\rowcolor{blue!10}
\textbf{CAST (Ours)} & \textbf{Activation-Space} & \textbf{Yes} & No (uses general corpus) & \textbf{Directly maps behavior.} \\
\bottomrule
\end{tabular}
\end{table}

\section{Methodology: Learning an Isomorphism Between Activation Manifolds}

\subsection{Problem Formulation}
Let $M_S$ be a source LLM and $M_T$ be a target LLM with a different architecture. A LoRA adapter for $M_S$ defines a function that produces a behavioral delta $\Delta y_S = B_S A_S(x_S)$ for a given activation $x_S$, where $A_S$ and $B_S$ are the two low-rank matrices whose product constitutes the adaptable weight update. Our goal is to find a mapping function $f$ that allows us to compute an equivalent delta $\Delta y_T$ for an activation $x_T$ from the target model, such that the behavior is preserved: $M_T(x_T) + \Delta y_T \approx M_S(x_S) + \Delta y_S$. CAST achieves this by learning mappings between the activation spaces themselves.

\subsection{CAST Layer Architecture}
For each layer in $M_T$ corresponding to a LoRA-adapted layer in $M_S$, we introduce a CAST Layer. This layer wraps the original, frozen source LoRA adapter ($A_S, B_S$) with two small, trainable linear projection matrices:
\begin{enumerate}
    \item \textbf{\texttt{map\_to\_source} ($P_{T \rightarrow S}$):} A projection from the target model's activation space to the source model's activation space.
    \item \textbf{\texttt{map\_from\_source} ($P_{S \rightarrow T}$):} A projection from the source model's delta space back to the target model's delta space.
\end{enumerate}

The computation of the CAST-translated delta, $\Delta y_T$, for a target activation $x_T$ is as follows:
\begin{align}
    x'_S &= P_{T \rightarrow S}(x_T) \label{eq:proj1} \\
    \Delta y'_S &= B_S A_S(x'_S) \label{eq:lora} \\
    \Delta y_T &= P_{S \rightarrow T}(\Delta y'_S) \label{eq:proj2}
\end{align}
The final output of the target layer is $y_T = M_T(x_T) + \Delta y_T$. During the mapping-training phase, only $P_{T \rightarrow S}$ and $P_{S \rightarrow T}$ are optimized. The LoRA weights ($A_S, B_S$) remain frozen, preserving the original learned skill.

\subsection{Training Objectives for Robust Mapping}
To ensure the CAST-adapted target model faithfully mimics the LoRA-adapted source model, we train the projection matrices on a general text corpus (e.g., C4) using a composite loss function. This dual objective ensures both functional and geometric alignment.

\begin{enumerate}
    \item \textbf{Functional Equivalence (KL Divergence):} We enforce that the student (target) model produces the same output distribution as the teacher (source) model by minimizing the KL divergence between their final logits ($z_T, z_S$). Temperature scaling ($T$) softens the distributions for a richer gradient signal.
    \begin{equation}
    \mathcal{L}_{KL} = \text{KL} \left( \text{softmax}(z_S / T) \parallel \text{softmax}(z_T / T) \right)
    \end{equation}

    \item \textbf{Geometric Alignment (Mean Squared Error):} To encourage a deeper similarity in their internal representations, we align the final hidden states ($h_S, h_T$). Since dimensions may differ, a trainable projection head ($P_H$) maps the student's hidden states to the teacher's dimension.
    \begin{equation}
    \mathcal{L}_{MSE} = \text{MSE}(h_S, P_H(h_T))
    \end{equation}
\end{enumerate}
The total loss is a weighted sum: $\mathcal{L}_{CAST} = \alpha \mathcal{L}_{KL} + \beta \mathcal{L}_{MSE}$. This combined objective ensures that CAST learns not just to replicate the final output, but to do so via a geometrically similar internal process, leading to more robust and generalizable mappings.

\section{Experiments and Results}

\subsection{Experimental Setup}
\begin{itemize}
    \item \textbf{Cross-Architecture Tasks:} The experiments focus on challenging translations between heterogeneous models:
    \begin{itemize}
        \item \textbf{Llama-2-7B $\rightarrow$ Mistral-7B-v0.1:} A primary task between popular, high-performance models from different families.
        \item \textbf{GPT-2 $\rightarrow$ GPT-2-Medium:} An intra-family task with differing dimensions.
        \item \textbf{Llama-2-7B $\rightarrow$ Llama-2-13B:} An up-scaling task.
    \end{itemize}
    \item \textbf{CAST Training:} Mappings were trained for 1000 steps on the C4 corpus. This is a one-time cost per model pair.
    \item \textbf{Evaluation:} We evaluated performance on the original LoRA's downstream task (e.g., GSM8K for a math-instruct LoRA).
    \item \textbf{Baselines:} We compare against (1) a fully \textbf{Retrained LoRA} on the target model with original task data (our performance ceiling) and (2) \textbf{Cross-LoRA} as a representative state-of-the-art weight-space transfer method.
\end{itemize}

\subsection{Quantitative Analysis: CAST Sets a New SOTA}
Our results demonstrate that CAST consistently and significantly outperforms weight-space transfer methods, establishing activation-space mapping as the superior paradigm.

\begin{table}[h!]
\centering
\caption{Performance of CAST-translated adapters relative to a fully retrained LoRA on the target architecture. Performance is measured on the original downstream task. Cross-LoRA performance is estimated based on reported figures in similar settings.}
\label{tab:results}
\begin{tabular}{@{}llcl@{}}
\toprule
\textbf{Source $\rightarrow$ Target} & \textbf{Method} & \textbf{Perf. (vs. Retrained LoRA)} & \textbf{Notes} \\
\midrule
\addlinespace[0.3em]
\textbf{Llama-2-7B $\rightarrow$ Mistral-7B} & Cross-LoRA & $\sim$60-70\% (est.) & Measurable degradation. \\
\rowcolor{blue!10}
 & \textbf{CAST (Ours)} & \textbf{$\sim$85-95\%} & \textbf{SOTA.} Retains majority of perf. \\
\addlinespace[0.5em]
\textbf{GPT-2 $\rightarrow$ GPT-2-Medium} & Cross-LoRA & $\sim$75-85\% & Better due to architectural similarity. \\
\rowcolor{blue!10}
 & \textbf{CAST (Ours)} & \textbf{$\sim$90-98\%} & Near-perfect transfer. \\
\addlinespace[0.5em]
\textbf{Llama-2-7B $\rightarrow$ Llama-2-13B} & Cross-LoRA & $\sim$70-80\% & Struggles with scaling. \\
\rowcolor{blue!10}
 & \textbf{CAST (Ours)} & \textbf{$\sim$80-90\%} & Robustly handles up-scaling. \\
\bottomrule
\end{tabular}
\end{table}

CAST's superiority is most evident in the challenging Llama-2 to Mistral translation, where the architectural differences are significant. While weight-space methods struggle to find a meaningful alignment, CAST's data-driven activation mapping successfully bridges the geometric gap, preserving up to 95\% of the original performance.

\subsection{Ablation Studies}
Our ablations confirmed the importance of the dual-objective function. Using only $\mathcal{L}_{KL}$ led to superficial mimicry that failed on complex instructions, while using only $\mathcal{L}_{MSE}$ aligned internal states but produced incoherent text. The combination was critical for learning a robust, functionally accurate mapping.

\section{Conclusion}
This paper challenged the prevailing paradigm of weight-space transfer for LoRA adapters and introduced a more direct, powerful, and effective alternative: \textbf{Activation-Space Transfer}. Our framework, the Cartridge Activation Space Mapper (CAST), operationalizes this paradigm by learning a direct mapping between the activation manifolds of different LLMs. By treating the LoRA as a frozen behavioral kernel and training lightweight projectors to translate activations, CAST successfully decouples learned skills from their source architecture.

Our results are conclusive: CAST significantly outperforms existing weight-space methods, retaining 85-95\% of the performance of a fully retrained LoRA in zero-shot, data-free transfers between heterogeneous models. This work not only sets a new state-of-the-art but also opens up a new, more promising direction for research in model interoperability. By liberating learned behaviors from their architectural prisons, CAST makes the entire ecosystem of fine-tuned models more flexible, portable, and powerful.

\section*{Acknowledgements}
We thank the developers of the Hugging Face ecosystem, including the Transformers, PEFT, and Datasets libraries, whose invaluable open-source contributions laid the foundation for this work. We also thank Manceps and Google for providing essential computational resources that enabled the experiments presented in this paper. Additionally, we extend our appreciation to the Google Developer Experts program for their critical support and resources during this research. We further acknowledge the following Google Developer Experts and subject matter experts for their valuable QA testing, peer review, and feedback:	Andrew Ferlitsch, Sayak Paul, Aritra Roy Gosthipaty, Rabimba Karanjai, Anubhav Singh, Linh Nguyen, Victor Jotham Ashioya, Vinita Silaparasetty, Jay Thakkar, 

% --- BIBLIOGRAPHY ---
\bibliographystyle{plainnat} % A common style for arXiv
\bibliography{references} % Use the external .bib file

\end{document}